\documentclass[journal]{IEEEtran}
\usepackage{cite}
\usepackage{amsmath,amssymb,amsfonts}
\usepackage{algorithm,algpseudocode}
\usepackage{graphicx,color}
\usepackage{textcomp}
\usepackage{xcolor}
\usepackage{hyperref}
\hypersetup{hidelinks=true}
\def\BibTeX{{\rm B\kern-.05em{\sc i\kern-.025em b}\kern-.08em
    T\kern-.1667em\lower.7ex\hbox{E}\kern-.125emX}}
\AtBeginDocument{\definecolor{tmlcncolor}{cmyk}{0.93,0.59,0.15,0.02}\definecolor{NavyBlue}{RGB}{0,86,125}}

\usepackage{soul}
\usepackage{subcaption}
\usepackage{listings} 
\usepackage{courier} 
\usepackage{lipsum} 
\usepackage{tabularx}
\usepackage{array}
\usepackage{booktabs}

\usepackage{multirow}
\usepackage{multicol}
\usepackage{rotating}
\usepackage{orcidlink}
\usepackage{tikz}
\usetikzlibrary{arrows.meta, positioning, calc, shapes.geometric, fit, backgrounds, matrix, shapes, decorations.pathreplacing}
\usepackage{svg}
\usepackage{makecell}
\usepackage[table]{xcolor}
\usepackage{siunitx}
\usepackage{balance}

\usepackage{threeparttable}
\usepackage{tabularx}
\usepackage{adjustbox}
\usepackage{subcaption}
\usepackage{multirow}

\begin{document}
\title{Loosely Coupled Factor Graph Optimization for Pseudolite-Augmented Navigation
\thanks{This work was supported by the Federal Ministry for Economic Affairs and Energy, Germany, as part of the research project ``Firefly2” (FKZ-50NA2404B). This work has also been conceptually supported by the German Federal Ministry of Research, Technology, and Space (BMFTR) under the Robotics Institute Germany (RIG).}
}

\author{Chih-Chun~Chen, Lipeng~Tan, Shiyu~Bai, and~Heike~Vallery
}


\maketitle

\begin{abstract}
In Global Navigation Satellite System (GNSS)-degraded environments, pseudolites (PLs) provide additional signal sources to enhance positioning performance, but their integration in optimization-based frameworks remains limited. 
This paper presents a loosely coupled factor graph optimization (FGO) framework that fuses the GNSS/PL least-squares (LS) solutions with inertial measurement unit (IMU) data. The evaluation considers low GNSS visibility scenarios with four high-elevation GNSS satellites and up to two PL transmitters over an 80~s window. FGO achieves a 22.8\% to 41.3\% reduction in mean 3D error compared to standard LS methods. Compared to a GNSS-IMU baseline, incorporating PL transmitters further improves positioning accuracy, with performance depending on geometry.
\end{abstract}



\section{Introduction}

Accurate localization is essential for autonomous vehicles. The Global Navigation Satellite System (GNSS) provides absolute positioning but degrades in environments such as tunnels and urban canyons due to blockage, multipath, and signal outages \cite{zhu2018urban_integrity, liu2023pl_smartcity}. To improve robustness, GNSS is commonly fused with inertial measurement units (IMUs), which provide high-rate motion information but suffer from drift if not corrected frequently \cite{shin2005_low_cost_ins}. 

Pseudolites (PLs), or pseudo-satellites, transmit GNSS-like signals and can be flexibly deployed to improve signal availability and geometry in GNSS-degraded environments. Since PL signals share the same structure as GNSS measurements and are compatible with conventional GNSS receivers, they can be incorporated into existing navigation frameworks without requiring additional hardware.

Most existing GNSS/PL integration approaches rely on least-squares (LS) \cite{lu2024indoor, huang2024high}, and do not incorporate IMU data. In contrast, factor graph optimization (FGO) has gained attention for multi-sensor navigation due to its ability to incorporate historical measurements and perform iterative optimization \cite{factor_graphs_for_robot_perception, zhang_continuous-time_multi_2024, wen_FGOEKF_2021}. 

To the best of the authors’ knowledge, no existing work has integrated PL with other sensors using FGO. This work proposes a loosely coupled (LC) FGO framework that fuses GNSS, PL, and IMU measurements for navigation in GNSS-degraded environments. The approach is validated using real-world data with two PL transmitters.

\section{Methods} \label{sec: methods}

\subsection{State Variables and Factor Formulations}
The state variables at time $t_i$ are defined as:
\begin{equation}
    \boldsymbol{x}_i = 
    \left\{
    \boldsymbol{R}^e_{b,i}, \;
    \boldsymbol{p}^e_{b,i}, \;
    \boldsymbol{v}^e_{b,i}, \;
    \boldsymbol{b}_{a,i}, \;
    \boldsymbol{b}_{g,i}
    \right\},
\label{eq: state_variables}
\end{equation}
where $\boldsymbol{R}^e_{b} \in SO(3)$ denotes the rotation matrix from the body frame ($b$) to the Earth-centered, Earth-fixed frame (ECEF, $e$), $\boldsymbol{p}^e_{b}, \boldsymbol{v}^e_{b} \in \mathbb{R}^3$ are the position and velocity expressed in ECEF frame, $\boldsymbol{b}_{a}, \boldsymbol{b}_{g} \in \mathbb{R}^3$ are the accelerometer and gyroscope biases. Below are the factors considered in the graph, where $\boldsymbol{\Sigma}^{\mathrm{imu}}$, $\boldsymbol{\Sigma}^{ba}$, $\boldsymbol{\Sigma}^{bg}$, and $\boldsymbol{\Sigma}^{P}$ denote the covariances of IMU preintegration, accelerometer bias, gyroscope bias, and GNSS/PL LS position, respectively.

\begin{itemize}
    \item \textbf{Preintegrated IMU Factor:} The motion between two consecutive states at timestamps $t_i$ and $t_j$ is modeled using the IMU preintegration formulation of \cite{forster2015imu}. The error function is defined using a residual vector that includes rotation $\boldsymbol{r}_{\Delta \boldsymbol{R}_{ij}}$, velocity $\boldsymbol{r}_{\Delta \boldsymbol{v}_{ij}}$, and position errors $\boldsymbol{r}_{\Delta \boldsymbol{p}_{ij}}$: 
    \begin{align}
        \|\boldsymbol{e}_{ij}^{\mathrm{imu}}\|^2
        &=
        \left\|
        \begin{bmatrix}
        \boldsymbol{r}_{\Delta \boldsymbol{R}_{ij}}^{T} &
        \boldsymbol{r}_{\Delta \boldsymbol{v}_{ij}}^{T} &
        \boldsymbol{r}_{\Delta \boldsymbol{p}_{ij}}^{T}
        \end{bmatrix}^{T}
        \right\|^2_{\boldsymbol{\Sigma}^{\mathrm{imu}}} \,.
        \label{eq: imu_factor}
    \end{align}
    
    \item \textbf{IMU Bias Factor:} 
    The biases error is modeled as in \cite{forster2015imu}:
    \begin{align}
        \|\boldsymbol{e}_{ij}^{\mathrm{b}}\|^2
        =
        \|\boldsymbol{b}_{a,j}-\boldsymbol{b}_{a,i}\|^2_{\boldsymbol{\Sigma}^{ba}}
        +
        \|\boldsymbol{b}_{g,j}-\boldsymbol{b}_{g,i}\|^2_{\boldsymbol{\Sigma}^{bg}} \,.
    \end{align}
    
    \item \textbf{GNSS/PL-Position Factor:} The observed antenna position $\tilde{\boldsymbol{p}}_{\text{ant}}^{e}$ is obtained via LS using combined GNSS and PL pseudorange measurements, since PL signals have the same structure as GNSS. The estimated antenna position is calculated as $\boldsymbol{p}_{\text{ant}, i}^{e} = \boldsymbol{p}_{b, i}^{e} + \boldsymbol{R}_{b, i}^{e} \boldsymbol{l}_{\text{ant}}^{b}$, where $\boldsymbol{l}_{\text{ant}}^{b}$ is the lever arm from the antenna to the IMU center. The error function is defined as:
    \begin{equation}
        \left\| \boldsymbol{e}_{i}^{\mathrm{p}} \right\|^2 = \left\| \boldsymbol{p}_{\text{ant}, i}^{e} - \tilde{\boldsymbol{p}}_{\text{ant}, i}^{e} \right\|_{\boldsymbol{\Sigma}^{P}}^2 \, .
        \label{eq: gnss_pl_position_factor}
    \end{equation}
\end{itemize}

\subsection{Factor Graph Structure}

\begin{figure}[t]
    \centering
    \resizebox{0.9\linewidth}{!}{%
    \begin{tikzpicture}[
            var/.style={circle,draw,minimum size=9mm,inner sep=1pt,fill=white,font=\small,align=center},
            fact/.style={rectangle,draw,minimum width=10mm,minimum height=6mm,inner sep=2pt,fill=black!08,font=\small},
            fact2/.style={rectangle,draw,minimum width=8mm,minimum height=4mm,inner sep=2pt,fill=black!08,font=\small},
            ghost/.style={circle,draw,densely dashed,minimum size=9mm,inner sep=1pt,fill=white!90,font=\small,opacity=0.6},
            factor edge/.style={thick}
        ]

        \node[var] (x0) at (0,0) {$\boldsymbol{x}_0$};
        \node[var] (x1) at (3,0) {$\boldsymbol{x}_1$};
        \node[var] (x2) at (6,0) {$\boldsymbol{x}_2$};

        \node[fact] (g0) at (0,1.5) {GNSS/PL};
        \node[fact] (g1) at (3,1.5) {GNSS/PL};
        \node[fact] (g2) at (6,1.5) {GNSS/PL};

        \node[fact] (u01) at (1.5, 1.0) {IMU};
        \node[fact] (u12) at (4.5, 1.0) {IMU};

        \node[fact] (ib0) at (1.5, 0) {IMU bias};
        \node[fact] (ib1) at (4.5, 0) {IMU bias};

        \node[fact] (prior) at (-1.0,0.6) {prior};

        \draw[factor edge] (g0) -- (x0);
        \draw[factor edge] (g1) -- (x1);
        \draw[factor edge] (g2) -- (x2);

        \draw[factor edge] (prior) -- (x0);

        \draw[factor edge] (u01.west) -- (x0);
        \draw[factor edge] (u01.east) -- (x1);
        \draw[factor edge] (u12.west) -- (x1);
        \draw[factor edge] (u12.east) -- (x2);

        \draw[factor edge] (ib0.west) -- (x0);
        \draw[factor edge] (ib0.east) -- (x1);
        \draw[factor edge] (ib1.west) -- (x1);
        \draw[factor edge] (ib1.east) -- (x2);


        \node[ghost] (xghost) at (8,0) {...};
        \node[fact2,opacity=0.35] (gghost) at (8,1.5) {...};
        \node[fact2,opacity=0.35] (ughost) at (7,0.0) {...};
        \node[fact2,opacity=0.35] (ighost) at (7,0.8) {...};

        \draw[factor edge,gray,densely dashed,opacity=0.35] (gghost) -- (xghost);
        \draw[factor edge,gray,densely dashed,opacity=0.35] (ughost.west) -- (x2);
        \draw[factor edge,gray,densely dashed,opacity=0.35] (ughost.east) -- (xghost);
        \draw[factor edge,gray,densely dashed,opacity=0.35] (ighost.west) -- (x2);
        \draw[factor edge,gray,densely dashed,opacity=0.35] (ighost.east) -- (xghost);

    \end{tikzpicture}%
    }
    \caption{Overview of the factor graph.} 
    \label{fig:factor_graph_structure}
\end{figure}
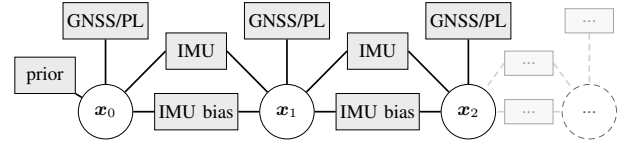

Fig.~\ref{fig:factor_graph_structure} illustrates the overall structure of the proposed factor graph. As new GNSS/PL measurements arrive, corresponding state nodes and measurement factors are appended to the factor graph, such that the number of state nodes is determined by the GNSS/PL measurement epochs. Between consecutive GNSS/PL measurements, high-rate IMU measurements are preintegrated to form a single IMU factor connecting the associated states. Additionally, a prior factor is introduced at the initial epoch to constrain the initial state.

The optimization problem is formulated as:
\begin{align}
    \hat{\boldsymbol{x}} = \arg\min_{\boldsymbol{x}} \biggl(&
    \left\lVert \boldsymbol{e}^{0} \right\rVert^{2}
    + \sum_{i=1}^{K} \left\lVert \boldsymbol{e}^{\mathrm{p}}_{i} \right\rVert^{2} \nonumber\\
    &\quad+ \sum_{i=1}^{K-1} \left( \left\lVert \boldsymbol{e}^{\mathrm{imu}}_{i,i+1} \right\rVert^{2} 
    + \left\lVert \boldsymbol{e}^{\mathrm{b}}_{i,i+1} \right\rVert^{2} \right) \biggl) \,,
\end{align}
where the error term $\boldsymbol{e}^{0}$ represents the prior factor. The index notation $K$ is the number of state nodes, which correspond to the GNSS/PL measurement epochs.

\subsection{Implementation and Measurement Setups} \label{sec: impl_meas}

The measurement campaign was conducted in a parking lot at the University of the Bundeswehr Munich, involving two PL transmitters and a static base station for single-differenced pseudorange correction (Fig.~\ref{fig: traj}). PL transmitters broadcast Galileo E1BC-like signals, and their hardware and time synchronization framework are described in \cite{hameed2024low, sanroma2024ejn, hameed2025plans}. A MEMS IMU and a multi-band GNSS/PL antenna were mounted on the rover roof rack, with the antenna positioned directly above the IMU (lever arm $\boldsymbol{l}^{b}_{\text{ant}} = [0;0;-0.1249]^T \,\mathrm{m}$).

Onboard devices included a software-defined radio (SDR) for GNSS/PL signal processing, an industrial computer for data logging, and a NovAtel receiver with the German SAPOS service to provide reference. The static base station consisted of an SDR and an antenna. The FGO framework was implemented based on the open-source \texttt{GTSAM} library \cite{gtsam}.

To emulate a GNSS-degraded scenario, only the four highest-elevation GPS satellites were used, with cases including zero, one, or two PL transmitters. The total evaluation duration was \SI{80}{s}.

\section{Results and Discussion} \label{sec: results}

\begin{figure}
    \centering
    \includegraphics[width=1.0\linewidth]{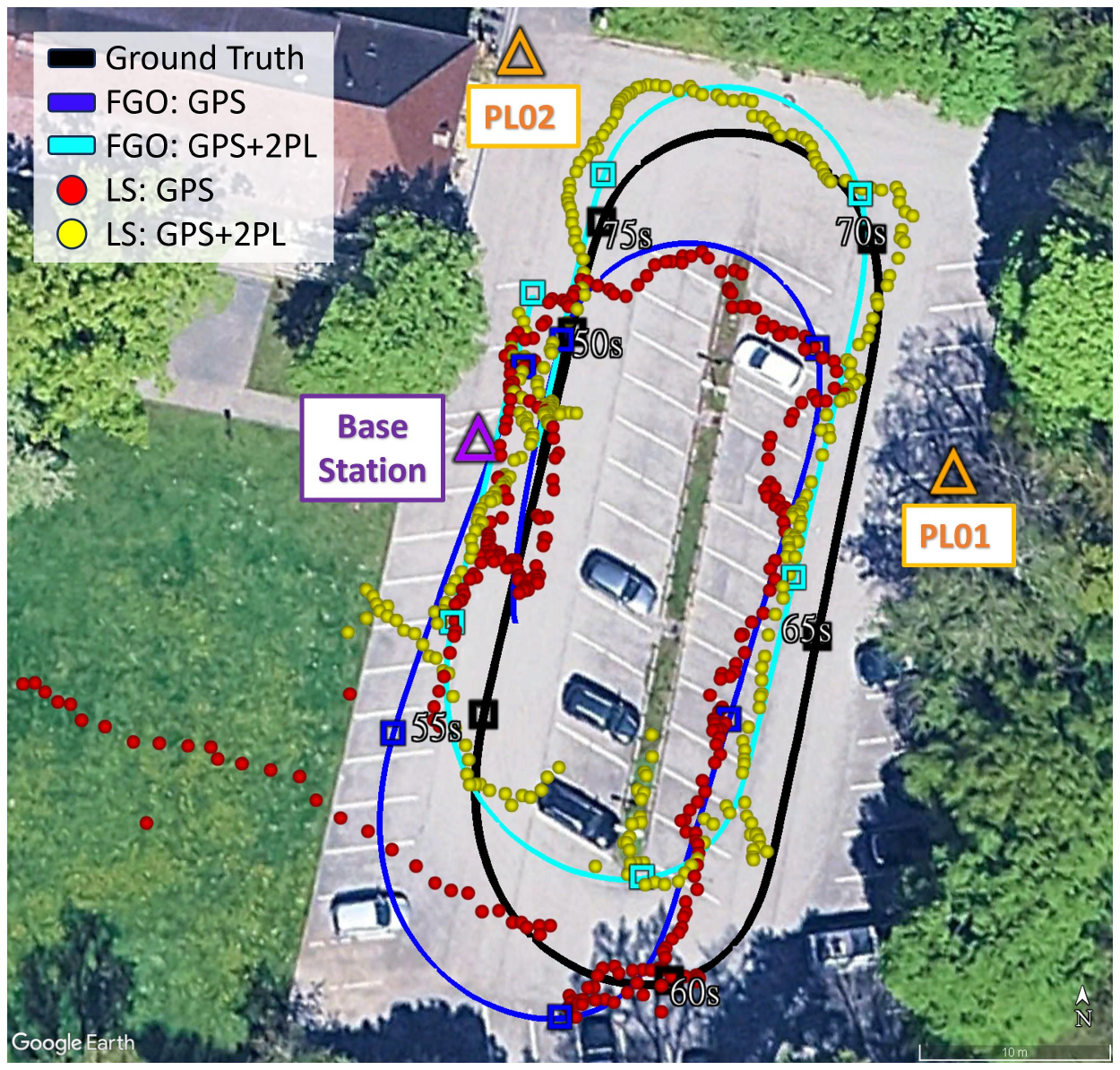}
    \caption{Partial trajectory results (\SI{50}{s}-\SI{80}{s}) for the GPS-only and GPS+2PL cases using LS and FGO, compared with ground truth. For readability, only the final \SI{30}{s} of the experiment are shown.}
    \label{fig: traj}
\end{figure}

\begin{table}[t]
\centering
\caption{Mean positional, horizontal, and vertical dilution of precision (PDOP, HDOP, VDOP), mean absolute positioning error (MAE), and maximum positioning error statistics. FGO includes IMU integration, whereas LS is IMU-free.}
\label{tab:combined_results}
\renewcommand{\arraystretch}{1.1}
\setlength{\tabcolsep}{3pt}
\resizebox{\columnwidth}{!}{%
\begin{tabular}{l ccc c cc cc}
\hline
\hline
\multirow{2}{*}{Signals} &
\multicolumn{3}{c}{Mean DOP} &
\multirow{2}{*}{Alg} &
\multicolumn{2}{c}{2D Error (m)} &
\multicolumn{2}{c}{3D Error (m)} \\
\cline{2-4} \cline{6-7} \cline{8-9}
& PDOP & HDOP & VDOP &
& MAE & Max & MAE & Max \\
\hline

\multirow{2}{*}{GPS}
& \multirow{2}{*}{8.75} & \multirow{2}{*}{6.07} & \multirow{2}{*}{6.31}
& LS  & 5.20 & 24.84 & 15.21 & 103.6 \\
& & &
& FGO & 4.58 & 8.79  & 8.93  & 31.11 \\
\hline

\multirow{2}{*}{GPS + 2PL}
& \multirow{2}{*}{3.11} & \multirow{2}{*}{1.54} & \multirow{2}{*}{2.50}
& LS  & 4.63 & 12.82 & 5.99 & 20.55 \\
& & &
& FGO & 3.73 & 6.94  & 3.94 & 6.94 \\
\hline

\multirow{2}{*}{GPS + PL01}
& \multirow{2}{*}{4.02} & \multirow{2}{*}{2.60} & \multirow{2}{*}{2.63}
& LS  & 4.75 & 12.01 & 6.54 & 16.16 \\
& & &
& FGO & 3.99 & 9.18  & 4.68 & 11.92 \\
\hline

\multirow{2}{*}{GPS + PL02}
& \multirow{2}{*}{4.35} & \multirow{2}{*}{1.74} & \multirow{2}{*}{3.96}
& LS  & 3.99 & 15.63 & 6.23 & 18.18 \\
& & &
& FGO & 3.15 & 7.89  & 4.81 & 8.83 \\
\hline
\hline
\end{tabular}%
}
\end{table}

Across all configurations, FGO consistently outperforms LS (Table~\ref{tab:combined_results}), which is also reflected in Fig.~\ref{fig: traj}. The MAE 2D improvement is 11.9\%-21.1\%, and the MAE 3D improvement is 22.8\%-41.3\%. The largest gain appears in the GPS-only setting. Compared to LS, FGO produces smoother trajectories and is less sensitive to measurement noise. This demonstrates the robustness of FGO-based sensor fusion under limited observability.

Positioning performance varies with the signal configuration and is not solely determined by the number of transmitters. The GPS-only case shows the largest error, while PL inclusion improves accuracy and reduces lateral bias in the trajectories (Fig.~\ref{fig: traj}), likely due to satellite measurement errors. The GPS+PL02 case achieves the lowest 2D error, whereas the GPS+2PL provides the best 3D performance. This indicates that the impact of additional transmitters depends on their spatial contribution and differs between horizontal and vertical accuracy.

The DOP trends align with the observed positioning performance. The GPS+2PL configuration has the lowest PDOP and achieves the best 3D accuracy. Compared to single-PL cases, GPS+PL02 shows better HDOP but worse VDOP and PDOP than GPS+PL01, which matches the 2D and 3D performance differences between the two. While DOP explains the overall trend, positioning errors are also influenced by multipath, clock bias, and measurement noise. Therefore, DOP alone does not fully determine accuracy, as illustrated by the GPS+2PL case, which has the best HDOP but not the best 2D performance.

\section{Conclusion} \label{sec: conclusion}

Experimental results show that the proposed loosely coupled FGO framework, integrating GNSS, PL, and IMU data, consistently outperforms traditional least squares solutions across all signal configurations. The inclusion of PL signals further enhanced accuracy, although the benefit depends on the transmitters' geometry. These findings validate the effectiveness of FGO for PL-assisted navigation in limited-satellite environments. Future work will investigate tightly coupled integration to further improve robustness and accuracy.

\newpage
\balance
\bibliographystyle{IEEEtran} 
\bibliography{references}

%








\end{document}